\documentclass[11pt]{article}
\usepackage[preprint]{acl}

\usepackage{times}
\usepackage{latexsym}

\usepackage[T1]{fontenc}

\usepackage[utf8]{inputenc}

\usepackage{microtype}

\usepackage{inconsolata}

\usepackage{graphicx}
\usepackage{tcolorbox}
\newcommand{\eat}[1]{}

\usepackage{latexsym}
\usepackage{amsfonts}
\usepackage{amsmath}
\usepackage{xcolor}
\usepackage{colortbl}
\usepackage{epsfig}
\usepackage{xspace}
\usepackage{graphicx}
\usepackage{subfigure}
\usepackage{paralist}
\usepackage{enumerate}
\usepackage[color,matrix,arrow,all]{xy}
\usepackage{comment}
\usepackage{booktabs}
\usepackage{balance}
\usepackage{stmaryrd}
\usepackage{pifont}
\usepackage{hhline}
\usepackage{listings}
\usepackage{array}
\usepackage{float}
\usepackage[flushleft]{threeparttable}

\usepackage{mathrsfs}
\usepackage{makecell}
\usepackage{xparse}
\usepackage{wrapfig}

\usepackage{epsfig}
\usepackage{multirow}
\usepackage{url}

\usepackage{multirow}
\usepackage{natbib}
\usepackage{graphicx}

\usepackage{listings}
\usepackage{framed}
\usepackage{xcolor}
\usepackage{color}
\usepackage{geometry}
\usepackage{changepage}
\colorlet{shadecolor}{gray!20}

\definecolor{shadecolor}{RGB}{220,220,220}

\definecolor{inputcolor}{RGB}{255,139,35}
\definecolor{outputcolor}{RGB}{120,212,252}
\definecolor{embedcolor}{RGB}{254,127,156}
\definecolor{maskcolor}{RGB}{122,128,255}
\definecolor{ecolor}{RGB}{58,149,54}

\definecolor{highcolor}{RGB}{255,153,153}
\definecolor{midcolor}{RGB}{255,204,204}
\definecolor{lowcolor}{RGB}{204,229,255}

\usepackage{tikz}
\usetikzlibrary{shapes,decorations}
\usetikzlibrary{calc}

\usepackage[export]{adjustbox}

\definecolor{green}{RGB}{0,128,0}

\definecolor{yellow}{RGB}{255,200,18}

\newcommand{\bi}{\begin{itemize}}
\newcommand{\ei}{\end{itemize}}

\newcommand{\be}{\begin{enumerate}}
\newcommand{\ee}{\end{enumerate}}
\newcommand{\beqn}{\begin{eqnarray*}}
\newcommand{\eeqn}{\end{eqnarray*}}

\newcommand{\eg}{{\em e.g.,}\xspace}

\makeatletter
    \newcommand\figcaption{\def\@captype{figure}\caption}
    \newcommand\tabcaption{\def\@captype{table}\caption}
\makeatother

\tikzstyle{mybox} = [draw=black, fill=black!5, thick,
   rectangle, rounded corners, inner sep=0pt, inner ysep=6pt]
\tikzstyle{fancytitle} =[fill=black, text=white]

\NewDocumentCommand{\nan}{ mO{} }{\textcolor{blue}{\textsuperscript{\textit{Nan}}\textsf{\textbf{\small[#1]}}}}

\NewDocumentCommand{\yang}{ mO{} }{\textcolor{green}{\textsuperscript{\textit{yang}}\textsf{\textbf{\small[#1]}}}}

\NewDocumentCommand{\zzx}{ mO{} }{\textcolor{yellow}{\textsuperscript{\textit{zzx}}\textsf{\textbf{\small[#1]}}}}

\usepackage{amsmath}
\usepackage{graphicx}
\usepackage{times}
\usepackage{amssymb}
\usepackage{multirow}

\usepackage[T1]{fontenc}

\usepackage[utf8]{inputenc}

\usepackage{microtype}

\usepackage{inconsolata}

\usepackage[ruled,linesnumbered]{algorithm2e}
\usepackage{booktabs} 
\usepackage{siunitx} 

\newcommand{\qa}{{MDMEQA}\xspace}
\newcommand{\sys}{{DocSage}\xspace}
\usepackage{multirow}
\usepackage{mdframed}
\title{DocSage: An Information Structuring Agent for Multi-Doc Multi-Entity Question Answering}

\author{
 \textbf{Teng Lin\textsuperscript{1}},
 \textbf{Yizhang Zhu\textsuperscript{1}},
 \textbf{Zhengxuan Zhang\textsuperscript{1}},
 \textbf{Yuyu Luo\textsuperscript{1,2}},
 \textbf{Nan Tang\textsuperscript{1,2}\thanks{Nan Tang is the corresponding author.}}
 \\
  \textsuperscript{1}The Hong Kong University of Science and Technology (Guangzhou)
  \\
 \textsuperscript{2}The Hong Kong University of Science and Technology
 \\
 \small{
    \href{mailto:email@domain}{\{tlin280,zzhang393,yzhu305\}@connect.hkust-gz.edu.cn}
 }
 \small{
    \href{mailto:email@domain}{\{yuyuluo,nantang\}@hkust-gz.edu.cn}
 }
 \\
}

\begin{document}
\maketitle

\begin{abstract}
Multi-document Multi-entity Question Answering (MDMEQA) inherently demands models to track implicit logic between multiple entities across scattered documents. However, existing Large Language Models (LLMs) and Retrieval-Augmented Generation (RAG) frameworks suffer from critical limitations: standard RAG’s vector similarity-based coarse-grained retrieval often omits crucial facts, graph-based RAG fails to efficiently integrate fragmented complex relationship networks, and both lack schema awareness, leading to inadequate cross-document evidence chain construction and inaccurate entity relationship deduction. To address these challenges, we propose \textbf{DocSage}, an end-to-end agentic framework that integrates dynamic schema discovery, structured information extraction, and schema-aware relational reasoning with error guaranties. DocSage operates through three core modules: (1) A schema discovery module dynamically infers query-specific minimal joinable schemas to capture essential entities and relationships; (2) An extraction module transforms unstructured text into semantically coherent relational tables, enhanced by error-aware correction mechanisms to reduce extraction errors; (3) A reasoning module performs multi-hop relational reasoning over structured tables, leveraging schema awareness to efficiently align cross-document entities and aggregate evidence. This agentic design offers three key advantages: precise fact localization via SQL-powered indexing, natural support for cross-document entity joins through relational tables, and mitigated LLM attention diffusion via structured representation. Evaluations on two MDMEQA benchmarks demonstrate that DocSage significantly outperforms state-of-the-art long-context LLMs and RAG systems, achieving more than 27\% accuracy improvements respectively. Our findings validate that the structured data representation and agentic design  effectively enhances complex reasoning and answer precision for MDMEQA by addressing the fragmentation and schema scarcity of unstructured multi-document data.
The source code and data have been made available at https://anonymous.4open.science/r/DocSage-07A7.

\end{abstract}

\section{Introduction}

Multi-document Multi-entity Question Answering (\qa) has emerged as a cornerstone task in knowledge-intensive natural language processing, addressing scenarios where answers depend on connecting implicit logic across scattered entities in multiple unstructured documents. From clinical research, where understanding drug-disease relationships requires synthesizing data across hundreds of trial reports, to financial analysis that demands comparing corporate performance indicators from diverse filings, and legal practice that relies on cross-referencing clauses across contracts, \qa enables decision-making in high-stakes domains by unlocking insights from fragmented information~\citep{lin-etal-2025-mebench,wang2024loong}. At its core, \qa requires models to not only locate relevant facts but also to establish reliable cross-document entity alignments and construct coherent evidence chains, capabilities that directly determine the accuracy and trustworthiness of results.

Despite the growing importance of \qa, existing approaches face fundamental limitations that hinder their effectiveness. While powerful for single-document reasoning, LLMs struggle with multi-document scenarios due to context window constraints and attention diffusion, often failing to track all entity relationships across disjoint text segments~\citep{lin-etal-2025-mebench}. RAG frameworks are designed to augment LLMs with external knowledge, offering partial solutions but introducing new challenges: standard RAG relies on vector similarity for retrieval, leading to coarse-grained selection that prioritizes semantic overlap over entity relevance, often omitting critical facts necessary for cross-document reasoning~\citep{lin2025srag,DBLP:journals/corr/abs-2504-10036}. Graph-based RAG variants~\citep{edge2024local}, which model entity relationships as triples, improve multi-hop reasoning but struggle to efficiently integrate complex, fragmented relationship networks across documents; graph construction becomes computationally prohibitive as document count scales. A unifying flaw across these methods is the lack of schema awareness: without explicit structured representations tailored to the query, they cannot systematically organize scattered entities and relationships, resulting in disjointed evidence chains and inaccurate entity deduction.

To address these limitations, we propose \sys, an end-to-end information structuring agent designed specifically for \qa. Inspired by cognitive theories that humans transform raw information into structured knowledge to simplify complex reasoning~\citep{li2024structrag}, DocSage embodies an agentic paradigm that autonomously discovers schemas, structures unstructured text, and performs schema-aware reasoning while guaranteeing error control. The framework’s core innovation lies in its integration of dynamic structured representation with agentic decision-making, enabling it to tackle the fragmentation and schema scarcity of multi-document data head-on.

\sys operates through three interdependent core modules, each addressing a critical bottleneck in \qa: (1) The \textbf{Schema Discovery Module} dynamically infers query-specific minimal joinable schemas, identifying essential entities, attributes, and relationships required to answer the query, even if these elements are not explicitly mentioned in the input text. This ensures the framework focuses only on relevant information, avoiding the noise of over-parameterized structures. (2) The \textbf{Extraction Module} transforms unstructured document content into semantically coherent relational tables, augmented by error-aware correction mechanisms that quantify extraction uncertainty and rectify low-confidence outputs. This module mitigates the ambiguity of natural language and LLM stochasticity, reducing extraction errors that plague unstructured retrieval. (3) The \textbf{Reasoning Module} leverages schema awareness to perform multi-hop relational reasoning over structured tables, efficiently aligning entities across documents via joinable schema attributes and aggregating evidence to generate accurate answers. This structured reasoning paradigm eliminates attention diffusion, enabling precise tracking of entity relationships even across large document collections.

Together, these modules deliver \textbf{three key advantages}: (1) SQL-powered indexing enables pinpoint fact localization, avoiding the omission of critical entities; (2) relational tables natively support cross-document entity joins, simplifying evidence chain construction; and (3) structured representation mitigates LLM attention dilution, enhancing reasoning over fragmented data. Our evaluations on two representative \qa benchmarks~\citep{lin-etal-2025-mebench,wang2024loong} demonstrate that \sys outperforms state-of-the-art long-context LLMs and RAG systems by 27.2\% and 27\% in accuracy, respectively. These results validate that integrating schema-aware structuring with an agentic framework is a viable solution to \qa’s core challenges.

\section{Related Work}

\subsection{Retrieval mechanisms with LLMs}
The integration of retrieval mechanisms with large language models has been a cornerstone in advancing open-domain question answering (QA). Early RAG frameworks, pioneered by ~\citep{lewis2020retrieval}, demonstrated the value of combining dense passage retrieval with generative models, but their efficacy diminishes in multi-entity scenarios where answers require synthesizing fragmented information across diverse documents. Subsequent refinements, such as REALM ~\citep{arora2023gar} and FiD ~\citep{izacard2021leveragin}, improved retrieval precision through cross-attention mechanisms, yet they inherently treat documents as isolated units, failing to model inter-entity relationships critical for questions like "Compare the research contributions of Turing Award winners in the last decade." While recent long-context LLMs (\eg Claude 3~\citep{claude3api2024}, GPT-4 Turbo~\citep{achiam2023gpt}) expand input windows to process hundreds of pages, empirical studies ~\citep{liu2025longcontext, lin2025Simplifying} reveal their tendency to ``overlook'' critical details in lengthy texts—a phenomenon termed contextual dilution—where key entities are lost due to attention saturation. Hybrid approaches, such as iterative retrieval with self-correction ~\citep{yoran2024making} and hierarchical summarization chains ~\citep{wang2023knowledgpt}, partially mitigate these issues but remain constrained by their linear processing of unstructured text, which obscures latent relationships between entities.

\subsection{Structure-Augmented Generation with LLMs}
Structured representation learning has emerged as a parallel strategy to enhance LLM reasoning. Methods like TableLLM ~\citep{zhang2025tablellm} pre-train models on tabular data to improve schema comprehension, while GraphRAG ~\citep{edge2024local} constructs knowledge graphs from retrieved snippets to enable relation-aware reasoning. However, these approaches either depend on pre-defined schemas—limiting adaptability to novel domains—or suffer from computational overhead when dynamically extracting entities from heterogeneous sources, which is similar in the case of Structed RAG~\citep{11107459}. Crucially, they treat structure creation as a post-retrieval step, decoupled from the initial information gathering process. In contrast, knowledge graph embedding techniques (\eg TransE~\citep{NIPS2013_1cecc7a7}) and template-based table generation prioritize static knowledge bases, rendering them ineffective for open-domain QA over evolving corpora like Wikipedia. 

\section{DocSage Framework}

\begin{figure*}[t!]
\begin{center}
\includegraphics[width=1\linewidth]{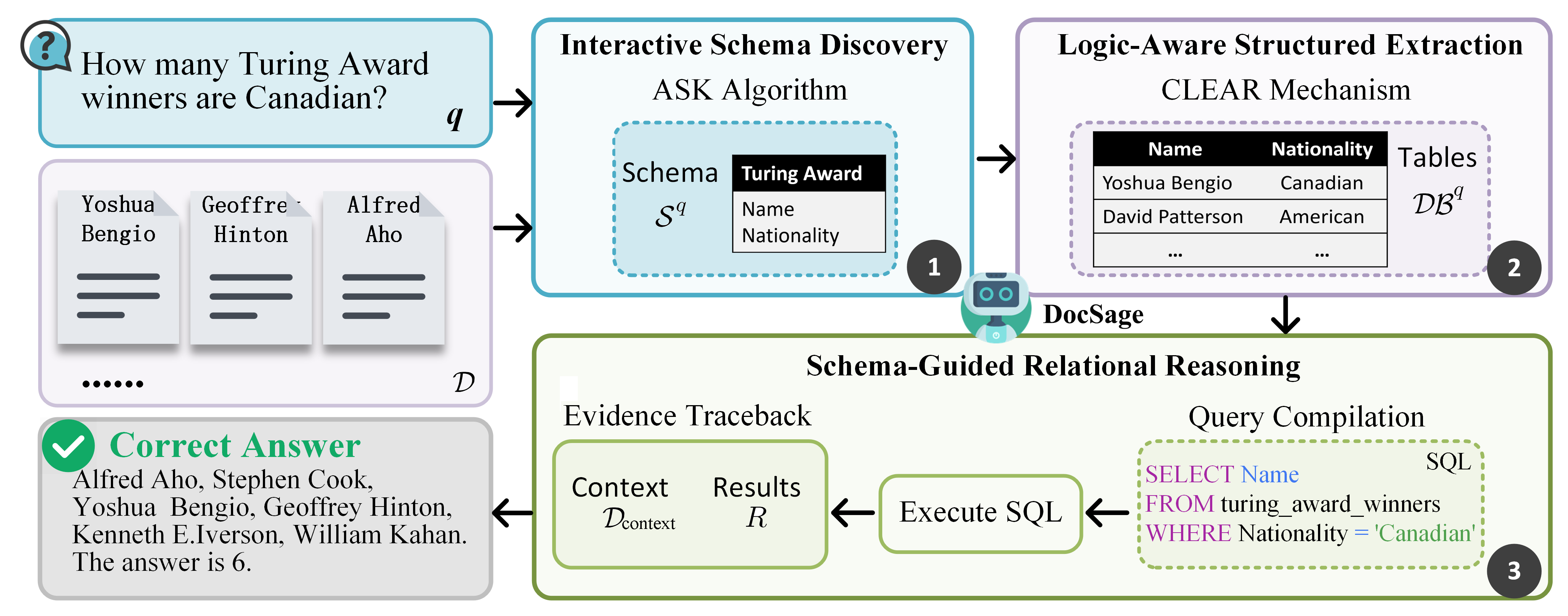}
\end{center}
  \caption{The overview of \sys framework, which consists of three core modules: Interactive Schema Discovery, Logic-Aware Structured Extraction, and Schema-Guided Relational Reasoning.}
  \label{fig:pipeline}
\end{figure*}

The DocSage framework is designed as an end-to-end agentic pipeline that transforms an unstructured document set \( \mathcal{D} = \{d_1, d_2, ..., d_N\} \) and a natural language query \( q \) into a precise, evidence-backed answer. Its core philosophy is to dynamically impose a \textbf{query-aware, error-guaranteed} relational structure on the documents through an agentic workflow. The system operates through three core, sequentially integrated modules: \textbf{Interactive Schema Discovery}, \textbf{Logic-Aware Structured Extraction}, and \textbf{Schema-Guided Relational Reasoning}. The overall pipeline is depicted in Figure~\ref{fig:pipeline}.

\subsection{Interactive Schema Discovery Module}

The primary objective of this module is to infer a \textbf{minimal, joinable} relational schema \( \mathcal{S}^q \) that precisely captures the entities, attributes, and relationships required to answer query \( q \), without relying on any predefined schema. To tackle the challenges of information fragmentation and schema ambiguity across multiple documents, we propose the \textbf{ASK (Active Schema Discovery via Knowledge-seeking Queries)} algorithm.

\textbf{ASK Algorithm Overview:} Unlike traditional two-phase (general$\rightarrow$refined) passive scanning, ASK frames schema discovery as an \textbf{interactive dialogue process} with the document set. After an initial scan, the system actively generates clarification questions to resolve ambiguities, confirm critical relationships, or supplement missing schema information. The \textbf{Algorithm Steps:}

\begin{itemize}
    \item \textbf{Step 1: Initial Schema Hypothesis Generation.} The system first uses a schema-derivation LLM \( M_{\text{schema}} \) to analyze the query \( q \) and a document subset \( \mathcal{D}_{\text{sample}} \subset \mathcal{D} \), producing an initial, potentially incomplete schema hypothesis \( \mathcal{S}_0^q \).
    \begin{equation}
        \mathcal{S}_0^q = \mathcal{F}_{\text{init}}(q, \mathcal{D}_{\text{sample}})
    \end{equation}
    
    \item \textbf{Step 2: Schema Consistency Analysis \& Question Generation.} The system applies \( \mathcal{S}_0^q \) to more documents and identifies three types of \textbf{uncertainty signals}:
    \begin{itemize}
        \item \textbf{Entity Alignment Conflicts:} The same entity is assigned different attributes across documents, or linking relationships are ambiguous.
        \item \textbf{Attribute Value Distribution Anomalies:} The value distribution of an attribute exhibits extreme outliers or logical contradictions (e.g., `Age=180').
        \item \textbf{Missing Relationships:} The multi-hop relationship paths required by the query cannot be fully established within the current schema.
    \end{itemize}
    For each type of uncertainty, a question generator \( \mathcal{F}_{\text{ask}} \) formulates a specific natural language question \( p \) to seek clarification:
    \begin{equation}
        P = \{p_j\} = \mathcal{F}_{\text{ask}}(\mathcal{S}_k^q, \mathcal{D}, \text{Uncertainty Signals})
    \end{equation}
    
    \item \textbf{Step 3: Answer Retrieval \& Schema Iterative Update.} The question set \( P \) is used as new ``queries'' to guide \textbf{targeted retrieval} over the full set \( \mathcal{D} \). Based on the retrieved evidence, a schema update function \( \mathcal{F}_{\text{update}} \) revises, expands, or prunes the current schema:
    \begin{equation}
        \mathcal{S}_{k+1}^q = \mathcal{F}_{\text{update}}(\mathcal{S}_k^q, \mathcal{A}(P, \mathcal{D}))
    \end{equation}
    where \( \mathcal{A} \) denotes the answer retrieval process for questions \( P \).
    
    \item \textbf{Step 4: Convergence \& Output.} The algorithm terminates when the schema \( \mathcal{S}_k^q \) stabilizes across consecutive iterations (i.e., no new critical uncertainties are identified) or all generated queries are adequately answered. The final schema \( \mathcal{S}^q \) is output.
\end{itemize}

This interactive design enables DocSage to \textbf{actively derive} a more robust and accurate query-specific schema, laying a solid foundation for subsequent precise extraction and reasoning.

\subsection{Logic-Aware Structured Extraction Module}

Given the target schema \( \mathcal{S}^q \), this module aims to populate it with tuples extracted from \( \mathcal{D} \) with high fidelity. Building upon the foundational data extraction pipeline~\citep{chao2025relationaldeepdive}, we introduce a \textbf{CLEAR (Cross-record Logic Enforcement for Accuracy Reinforcement)} correction mechanism. This mechanism not only assesses the confidence of individual extractions but also enforces cross-record logical consistency constraints.

\textbf{Step 1: Base Tuple Extraction.} For each document chunk \( d_k \), we use an extraction small-scale LM \( M_{\text{ext}} \) and the schema \( \mathcal{S}^q \) to generate a candidate tuple \( \mathbf{t}_k \).

\textbf{Step 2: CLEAR Correction Mechanism.} The correction process operates at two levels:
\begin{itemize}
    \item \textbf{Level A: Single-Point Confidence Assessment (via LoRA-Enhanced Calibration):} Instead of using the raw LM's hidden states, we fine-tune a lightweight LoRA (Low-Rank Adaptation) adapter to the base extraction model \( M_{\text{ext}} \) for each target schema \( \mathcal{S}^q \). This adapter is trained on a small labeled dataset \( \mathcal{D}_{\text{l}} \), making the model output more attuned to the current schema and query. We then compute a \textbf{calibrated confidence score} \( \text{conf}(\mathbf{t}_k) \), which combines model probability with a Conformal Prediction threshold based on \( \mathcal{D}_{\text{l}} \).
    \begin{equation}
        \text{conf}(\mathbf{t}_k) = \text{Ca}(\text{Prob}_{M_{\text{ext}}+\text{LoRA}}(\mathbf{t}_k | d_k, \mathcal{S}^q), \mathcal{D}_{\text{l}})
    \end{equation}
    
    \item \textbf{Level B: Cross-Record Logical Consistency Checking:} This is the core innovation of CLEAR. We define a set of \textbf{schema-dependent logical constraints} \( \mathcal{C}(\mathcal{S}^q) \), such as:
    \begin{itemize}
        \item \textbf{Functional Dependencies:} 
        `Person\_ID' $\rightarrow$ `Date\_of\_Birth' (one ID maps to one birth date).
        \item \textbf{Temporal Constraints:} 
        `Admission\_Time' $<$ `Discharge\_Time'.
        \item \textbf{Numerical Ranges:} 
        `Age' within a known safe range.
        \item \textbf{Foreign Key Referential Integrity:} `Stock\_Price.Company\_Ticker' must exist in `Company.Company\_Ticker'.
    \end{itemize}
    The system stages all candidate tuples \( \{\mathbf{t}_k\} \) in a temporary database and runs a lightweight \textbf{constraint validation engine}(SLM) to detect the set of tuples \( \mathcal{V} \) that violate \( \mathcal{C}(\mathcal{S}^q) \).
\end{itemize}

\textbf{Correction Decision \& Execution:} For each tuple with low confidence \( (\text{conf}(\mathbf{t}_k) < \tau_{\text{low}}) \) or involved in a logic violation \( (\mathbf{t}_k \in \mathcal{V}) \), the system triggers a correction workflow. The correction strategy is dynamically selected based on the violation type:
\begin{itemize}
    \item For simple omissions or low confidence, a committee of more powerful LLMs is used for re-extraction.
    \item For complex logic conflicts (e.g., foreign key mismatches), the conflicting group \( \{\mathbf{t}_i, \mathbf{t}_j\} \) along with relevant source text is submitted to a \textbf{Verification \& Disambiguation Sub-module}. This module may perform deeper contextual analysis or initiate a \textbf{targeted backtracking retrieval} to find decisive evidence from the original documents.
\end{itemize}
Finally, corrected and verified tuples are inserted into the final, high-quality relational database \( \mathcal{DB}^q \).

\subsection{Schema-Guided Relational Reasoning Module}

The reasoning module executes the query \( q \) directly on the precisely constructed database \( \mathcal{DB}^q \) and its schema \( \mathcal{S}^q \). The explicit presence of the schema transforms complex multi-hop reasoning into deterministic database operations.

\textbf{Query Compilation \& Optimization:} A reasoning LLM \( M_{\text{reason}} \) compiles \( q \) into an SQL query \( Q_{\text{SQL}} \). Benefiting from the explicit join keys and relationship definitions in \( \mathcal{S}^q \), the compiler can generate \textbf{highly optimized join queries}. For conditions involving multi-level nesting or aggregation, the compiler leverages schema information to \textbf{push down filters} and choose the most efficient join order, significantly improving query efficiency.
\begin{equation}
    Q_{\text{SQL}}^{\text{optimized}} = \mathcal{F}_{\text{compile\_optimize}}(q, \mathcal{S}^q)
\end{equation}

\textbf{Evidence Traceback \& Answer Generation:} Executing \( Q_{\text{SQL}}^{\text{optimized}} \) yields a structured result set \( R \). The system \textbf{automatically traces} the provenance of each row in \( R \) back to its originating tuples in \( \mathcal{DB}^q \) and further maps them to specific locations in the original document chunks \( \mathcal{D}_{\text{context}} \). \( M_{\text{reason}} \) synthesizes the final natural language answer \( a \) based on \( R \) and the complete provenance chain \( \mathcal{E} \):
\begin{equation}
    a, \mathcal{E} = \mathcal{F}_{\text{synthesize}}(R, \mathcal{DB}^q, \mathcal{D}_{\text{context}})
\end{equation}
This design ensures the \textbf{verifiability} of the answer, as each claim can be traced back to reliable, logically validated raw data.

\section{Experiment} 

\begin{table*}[t!]
  \centering
  \renewcommand\arraystretch{1}
  \caption{
   Experimental results for MEBench. 
  }
  {\small
  \begin{tabular}{lcccc}
    \toprule
    \multirow{2}{*}{\textbf{Method}} 
        & \multicolumn{4}{c} {\textbf{Accuracy}}\\
        \cmidrule(lr){2-5}
        & Comparison &Statistics &Relationship &Overall\\
    \midrule
    \rowcolor{black!8} \multicolumn{5}{c} {\textbf{All sets}}\\
     \midrule
    GPT-4o & 0.262 &0.353 &0.407 &0.338 \\
    GPT-4o + RAG &0.696 &0.579 &0.593 &0.620 \\
    GraphRAG &0.618 &0.558 &0.593 &0.586\\
    StructRAG &0.678 &0.588 &0.573 &0.612\\
    \textbf{\sys (Ours)} & \textbf{0.934} &\textbf{0.908} &\textbf{0.812}	&\textbf{0.892} \\
    \midrule
    \rowcolor{black!8}\multicolumn{5}{c} {\textbf{Set1 (0-10)}}\\
     \midrule
    GPT-4o &0.467 &0.595 &0.571 &0.548\\
    GPT-4o + RAG &0.870 &0.690 &0.755 &0.764\\
    GraphRAG &0.774 &0.761 &0.694 &0.748\\
    StructRAG & 0.838 &0.773 &0.735 &0.784\\
    \textbf{\sys (Ours)} & \textbf{0.968} &\textbf{0.929} &\textbf{0.837}	&\textbf{0.918} \\
    \midrule
    \rowcolor{black!8}\multicolumn{5}{c} {\textbf{Set2 (11-100)}}\\
     \midrule
    GPT-4o & 0.388 &0.505 &0.525 &0.473 \\
    GPT-4o + RAG &0.777 &0.613 &0.667 &0.679\\
    GraphRAG &0.714 &0.589 &0.707 &0.659\\
    StructRAG &0.793 &0.601 &0.657 &0.676\\
    \textbf{\sys (Ours)} & \textbf{0.952} &\textbf{0.923} &\textbf{0.818}	&\textbf{0.906} \\
     \midrule
    \rowcolor{black!8}\multicolumn{5}{c} {\textbf{Set3 (>100)}}\\
    \midrule
    GPT-4o &0.153 &0.214 &0.306 &0.219 \\
    GPT-4o + RAG &0.508 &0.350 &0.413 &0.415 \\
    GraphRAG &0.450 &0.344 &0.417 &0.396 \\
    StructRAG &0.492 &0.374 &0.359 &0.406 \\
    \textbf{\sys (Ours)} & \textbf{0.946} &\textbf{0.884} &\textbf{0.791}	&\textbf{0.879} \\
    \bottomrule
  \end{tabular}
  }
  \label{tab:mebench}
\end{table*}

\subsection{Experiment Setup}
\textbf{Evaluation Datasets.}
We evaluate our proposed method on two challenging \qa benchmarks: MEBench~\citep{lin-etal-2025-mebench} and Loong~\citep{wang2024loong}. MEBench is a specialized benchmark for multi-entity QA, comprising 4,780 methodically crafted questions. These are systematically categorized into three primary types: Comparison, Statistics, and Relationship which aims to provide comprehensive coverage of diverse and realistic multi-entity reasoning scenarios. Loong includes four distinct reasoning tasks: Spotlight Locating, Comparison, Clustering, and Chain of Reasoning, across four increasing document length settings. This design specifically tests a model's ability to locate and connect relevant information as it becomes more dispersed throughout longer documents.

\textbf{Implementation Details.}
In \sys, GPT-4o~\citep{achiam2023gpt} and Qwen3~\citep{yang2025qwen3technicalreport} are employed as the primary LLMs. For small-scale language model of information extraction, we use Mistral-7B~\citep{jiang2023mistral}. In the Reasoning module, we used GPT-4o as the reasoning model.

\textbf{Baselines.}
We selected baseline methods from widely adopted and state-of-the-art approaches in \qa. Among proprietary large language models , we selected the widely recognized GPT-4o~\citep{achiam2023gpt}, which serves as a strong standalone generative baseline. To evaluate retrieval-augmented strategies, we incorporated the standard RAG framework~\citep{lewis2020retrieval}, which segments documents into short chunks and uses a retriever to select the most relevant segments based on the input question. These are then used as context for GPT-4 during answer generation. We also included two advanced structured retrieval methods: (1) GraphRAG~\citep{edge2024local}, which constructs a knowledge graph from extracted (head, relation, tail) triples and uses graph retrieval and reasoning to enhance answer generation. (2) StructRAG~\citep{li2024structrag}, a structure-aware framework that dynamically identifies suitable structured representations for a given task, reconstructs textual content into that format, and performs inference over the organized data. This selection enables a comprehensive comparison across plain LLMs, naive retrieval, and more sophisticated graph-based or structure-aware augmentation techniques.

\textbf{Evaluation Metrics.}
For the MEBench benchmark, we employ \emph{Accuracy} as the primary evaluation metric to measure performance on the tasks. 
Within the Statistics category—specifically for the sub-tasks of Variance Analysis, Correlation Analysis, and Distribution Compliance, we evaluate the correctness of the selected columns and methods. 
For the Loong benchmark, we adhere to the original evaluation protocol and utilize the official evaluation code. Performance is measured using a dual mechanism: LLMs are prompted to output a confidence score between 0 and 100, and final answers are also evaluated via Exact Match (EM) rate to ensure precise alignment with ground-truth responses.

\subsection{Main Results}


\begin{table*}[t!]
  \centering
  \renewcommand\arraystretch{1}
  \caption{
   Experimental results for Loong. 
  }
  {\small
  \begin{tabular}{lcccccccccc}
    \toprule
    \textbf{Method} &\multicolumn{2}{c} {\textbf{Spotlight Locating}} &\multicolumn{2}{c} {\textbf{Comparison}} &\multicolumn{2}{c} {\textbf{Clustering}} &\multicolumn{2}{c} {\textbf{Chain of Reason}} &\multicolumn{2}{c} {\textbf{Overall}}\\
    \midrule
    \rowcolor{black!8} \multicolumn{11}{c} {\textbf{All sets}}\\
     \midrule
    GPT-4o &76.79 &0.65 &50.98 &0.29 &45.04 &0.10 &57.46 &0.27 &54.17 &0.26 \\
    GPT-4o + RAG &64.04 &0.44 &41.85 &0.26 &35.37 &0.03 &41.62 &0.19 &43.05 &0.18 \\
    GraphRAG &22.49 &0.00 &22.91 &0.01 &37.52 &0.03 &45.34 &0.23 &33.44 &0.07\\
    StructRAG &68.07 &0.40 &63.36 &0.36 &60.71 &0.14 &53.27 &0.18 &60.56 &0.23\\
    \textbf{\sys (Ours)} &\textbf{85.06} &\textbf{0.84} &\textbf{73.28} &\textbf{0.43} &\textbf{64.43} &\textbf{0.42} &\textbf{73.67}&\textbf{0.57} &\textbf{68.29} &\textbf{0.53}\\
    \midrule
    \rowcolor{black!8}\multicolumn{11}{c} {\textbf{10K-50K Tokens}}\\
     \midrule
   GPT-4o &87.38 &0.83 &65.56 &0.34 &58.15 &0.24 &83.21 &0.56 &71.81 &0.45\\ 
   GPT-4o + RAG &50.57 &0.35 &44.08 &0.27 &37.58 &0.05 &53.41 &0.35 &45.65 &0.23 \\
    GraphRAG & 32.30 &0.02 &28.15 &0.03 &41.52 &0.14 &55.38 &0.44 &41.64 &0.18\\ 
    StructRAG &76.02 &0.48 &77.09 &0.48 &66.43 &0.23 &69.20 &0.35 &70.82 &0.36 \\
    \textbf{\sys (Ours)} &91.12 &0.94 &87.10 &0.58 &67.97 &0.45 &90.84 &0.70 &80.09 &0.62 \\
    \midrule
    \rowcolor{black!8}\multicolumn{11}{c} {\textbf{50K-100K Tokens}}\\
     \midrule
    GPT-4o &88.50 &0.73 &61.01 &0.41 &48.79 &0.11 &63.33 &0.35 &59.55 &0.30 \\
    GPT-4o + RAG &67.60 &0.47 &47.21 &0.32 &39.73 &0.05 &47.07 &0.22 &46.33 &0.19 \\
    GraphRAG &24.55 &0.00 &14.15 &0.00 &37.48 &0.00 &45.79 &0.12 &32.73 &0.03 \\
    StructRAG &69.36 &0.42 &64.98 &0.37 &62.63 &0.17 &55.79 &0.19 &62.17 &0.24 \\
    \textbf{\sys (Ours)} &88.79 &0.88 &75.68 &0.50 &63.19 &0.42 &71.23 &0.55 &70.76 &0.54 \\
     \midrule
    \rowcolor{black!8}\multicolumn{11}{c} {\textbf{100K-200K Tokens}}\\
    \midrule
    GPT-4o &76.34 &0.66 &43.25 &0.21 &39.47 &0.04 &45.96 &0.09 &47.89 &0.19 \\
    GPT-4o + RAG &75.16 &0.56 &43.04 &0.28 &33.44 &0.02 &39.53 &0.14 &44.73 &0.18 \\
    GraphRAG &16.15 &0.00 &27.95 &0.00 &43.35 &0.00 &44.20 &0.17 &33.95 &0.04 \\
    StructRAG &67.25 &0.43 &56.59 &0.34 &57.10 &0.10 &48.74 &0.13 &56.76 &0.21 \\
    \textbf{\sys (Ours)} &81.44 &0.80 &68.12 &0.30 &61.72 &0.41  &68.15 &0.52 &62.43 &0.48 \\
     \midrule
    \rowcolor{black!8}\multicolumn{11}{c} {\textbf{200K-250K Tokens}}\\
    \midrule
    GPT-4o &37.53 &0.19 &24.45 &0.08 &31.01 &0.00 &33.55 &0.07 &31.73 &0.07\\
    GPT-4o + RAG &53.21 &0.24 &24.85 &0.10 &27.05 &0.00 &17.97 &0.00 &29.58 &0.07\\
    GraphRAG &17.85 &0.00 &27.20 &0.00 &21.33 &0.01 &34.15 &0.34 &23.94 &0.05\\
    StructRAG &58.01 &0.19 &56.73 &0.26 &57.72 &0.00 &36.42 &0.05 &52.45 &0.10\\
    \textbf{\sys (Ours)} &72.86 &0.71 &61.65 &0.30 &70.34 &0.39 &69.85 &0.54 &60.52 &0.47\\
    \bottomrule
  \end{tabular}
  }
  \label{tab:loong}
\end{table*}

Table~\ref{tab:mebench} presents experimental results alongside overall accuracy on MEBench, and Table~\ref{tab:loong} shows LLM-judged scores and exact match rate in Loong benchmark. The left indicator represents the Avg Scores (0-100), and the right one represents the Perfect Rate (0-1).
The experimental results demonstrate that the proposed \sys achieves state-of-the-art performance across both the MEBench and Loong benchmarks, significantly outperforming all baseline models. The superiority of \sys is consistent across all question types, multi-entity reasoning tasks, and document length settings, highlighting its robustness and effectiveness in handling complex \qa scenarios.

\textbf{MEBench Results.}
MEBench tests a model's ability to reason over multiple entities through Comparison, Statistics, and Relationship questions.
\begin{itemize}
 \item {Performance of \sys}: \sys achieved a remarkable overall accuracy of 89.2\%, which is a substantial improvement over the next best method, GPT-4o + RAG (62.0\%), by 27.2 percentage points. This indicates a fundamental advancement in multi-entity reasoning capabilities.
 \item {Consistency Across Question Types}: The superiority of \sys is consistent across all question categories, with accuracies of 93.4\% (Comparison), 90.8\% (Statistics), and 81.2\% (Relationship). This suggests that the method's underlying architecture is well-suited for the distinct logical demands of each question type.
 \item {Robustness to Increasing Entity Density}: A key finding is \sys's exceptional robustness as the number of entities increases. While all methods see a performance drop from Set1 (0-10 entities) to Set3 (>100 entities), the decline for \sys is minimal (from 91.8\% to 87.9\%). In contrast, competitors like GPT-4o + RAG and StructRAG suffer severe degradation (e.g., GPT-4o + RAG drops from 76.4\% to 41.5\%). This demonstrates \sys's superior ability to locate and synthesize entity information from a large, dispersed set of documents, which is a critical requirement for real-world multi-document QA. 
\end{itemize} 

\textbf{Loong Results.}
Loong evaluates a model's capability for specific reasoning tasks under the challenge of increasing document length.
\begin{itemize}
 \item {Superior Overall Performance}: \sys achieves the highest Overall Avg Score (68.29) and, more importantly, a dramatically higher Perfect Rate (0.53) compared to all other methods. The Perfect Rate metric is a stringent indicator of how often a model produces a fully correct answer. \sys's 0.53 rate is more than double that of the next best model (GPT-4o at 0.26), highlighting its precision and reliability.
 \item {Task-specific Strengths}: \sys excels in tasks requiring precise information location and complex reasoning. It leads in Spotlight Locating (0.84) and Chain of Reasoning (0.57), demonstrating an unmatched ability to find key facts and perform multi-step inferences. It also performs strongly in Comparison and Clustering.
 \item {Handling Long Document Contexts}: The results across increasing token lengths confirm \sys's scalability. While all models struggle with the longest documents (200K-250K tokens), \sys's performance decline is the least severe. It maintains a strong Avg Score of 60.52 and a dominant Perfect Rate of 0.47 in the most challenging setting, whereas other models see their Perfect Rates drop to 0.10 or below. This proves that \sys's method for structuring information is crucial for managing the complexity and information dispersion inherent in long-context scenarios. 
\end{itemize} 
\subsection{Analysis of results}

The experimental evaluation on two challenging multi-document QA benchmarks leads to the following conclusions:
\begin{itemize}
 \item {Significant Advancement}: The proposed \sys method establishes a new state-of-the-art for multi-document question answering, significantly outperforming strong baselines.
 \item {Robustness}: The most notable advantage of \sys is its robustness to scale. Its performance remains consistently high even as the number of entities or the length of the context increases dramatically, a scenario where other models fail significantly.
 \item {Effective Reasoning}: \sys demonstrates superior capabilities across a diverse set of reasoning tasks, from simple fact location (Spotlight) to complex, multi-step reasoning chains. Its high performance on Loong indicates that it produces correct answers more reliably and completely.
\end{itemize} 
These results strongly validate the design principles of \sys, suggesting that its structured approach to organizing and reasoning over information from multiple documents is highly effective for tackling the challenges of real-world, large-scale question answering. The detailed ablation study results are in~\ref{Apx:ablation}.

\section{Conclusion}
In conclusion, this paper introduces \sys, a novel agentic framework that addresses the fundamental challenges of Multi-document Multi-entity Question Answering by transforming unstructured text corpora into a dynamic, query-specific relational representation. Through its three synergistic modules—interactive schema discovery, logic-aware structured extraction with statistical error guarantees, and schema-guided relational reasoning, \sys effectively overcomes the limitations of coarse-grained retrieval and schema-agnostic reasoning inherent in standard RAG and long-context LLM approaches. Our comprehensive evaluation on established MDMEQA benchmarks demonstrates \sys's superior performance, achieving accuracy improvements of over 27\% by enabling precise fact localization, seamless cross-document entity joins, and robust multi-hop deduction. These results validate the core thesis that an agentic workflow centered on dynamic structure induction is a powerful paradigm for enhancing complex reasoning over fragmented, schema-scarce document collections.

\section{Limitations}

While \sys demonstrates significant advancements in \qa, several limitations warrant consideration for future work. First, the framework's performance and efficiency are inherently coupled with the capabilities of the underlying foundation models used for schema discovery and extraction, and it incurs non-trivial computational cost due to its iterative, multi-stage agentic process. Second, its current design assumes a degree of semantic coherence and factual consistency within the source documents; performance may degrade with extremely noisy, contradictory, or domain-specialized texts that challenge general-purpose LLMs.

\bibliography{refs/custom}

\appendix


\clearpage
\appendix
\section{Appendix}
\label{sec:appendix}

\subsection{End-to-End Agentic Workflow}

The agent orchestrates the sequential execution of the above modules, forming a closed-loop workflow. The \textbf{Key Advantages} are as follows:

\begin{itemize}
    \item \textbf{Active Schema Discovery (ASK)} addresses schema scarcity and ambiguity, surpassing passive schema inference.
    \item \textbf{Logic-Aware Extraction (CLEAR)} provides stronger error guarantees by combining statistical confidence with symbolic logic constraints.
    \item \textbf{Schema-Guided Reasoning} offloads computational burden from the LLM to an efficient database engine, avoiding attention diffusion and ensuring deterministic, explainable reasoning.
\end{itemize}

This system design enables DocSage to fundamentally address the core challenges of \qa: information fragmentation, implicit relationships, and high accuracy requirements.

\subsection{Ablation Study Results}
\label{Apx:ablation}
\subsubsection{Ablation Study Results for DocSage on MEBench}

To validate the effectiveness of each core component within the DocSage framework, we conduct a systematic ablation study. The performance of the full DocSage model is compared against several ablated variants on the MEBench dataset, with results presented in Table~\ref{tab:ablation}.

\begin{table*}[t]
    \centering
    \caption{Results of the ablation study on the MEBench benchmark.}
    \label{tab:ablation}
    \begin{tabular}{lcccc}
        \toprule
        \textbf{Method} & \textbf{Comparison} & \textbf{Statistics} & \textbf{Relationship} & \textbf{Overall} \\
        \midrule
        \textbf{DocSage (Full)} & \textbf{0.934} & \textbf{0.908} & \textbf{0.812} & \textbf{0.892} \\
        w/o Structured Extraction & 0.752 & 0.687 & 0.635 & 0.691 \\
        w/o Schema-Guided Reasoning & 0.821 & 0.794 & 0.703 & 0.773 \\
        w/o Schema Discovery & 0.845 & 0.775 & 0.724 & 0.781 \\
        w/o CLEAR (Logic Checking) & 0.901 & 0.862 & 0.783 & 0.849 \\
        w/ Passive Schema Discovery & 0.916 & 0.879 & 0.795 & 0.863 \\
        \bottomrule
    \end{tabular}
\end{table*}

\paragraph{Result Analysis:}
\begin{enumerate}
    \item \textbf{Core Importance of Structured Extraction:} The \textbf{removal of the Structured Extraction Module (w/o Structured Extraction)} results in the most severe performance degradation (an overall accuracy drop of 20.1 percentage points). This strongly proves that accurately transforming unstructured text into high-fidelity, computable relational data is the cornerstone of the entire framework. Its effectiveness directly impacts the reliability of all subsequent reasoning.

    \item \textbf{Key Role of Schema-Guided Reasoning:} The \textbf{removal of Schema-Guided Reasoning (w/o Schema-Guided Reasoning)} causes a significant performance loss (a drop of 11.9 percentage points). This validates the necessity of offloading complex multi-hop logic, joins, and aggregation operations to a deterministic SQL engine. This design effectively avoids the issues of attention diffusion and hallucinations in large language models during long-range reasoning, and is key to ensuring the accuracy of answers to complex questions.

    \item \textbf{Importance of Dynamic Schema Discovery:} The \textbf{removal of the Schema Discovery Module (w/o Schema Discovery)} also has a clear negative impact (a drop of 11.1 percentage points), indicating that dynamically inferring a query-specific relational schema is crucial for effectively organizing and connecting fragmented entity information. However, given the availability of high-quality structured data, its importance is slightly secondary to the two aforementioned modules.

    \item \textbf{Gains from Error Correction and Active Discovery:} Both \textbf{disabling the logic check of CLEAR (w/o CLEAR)} and \textbf{adopting passive schema discovery (w/ Passive Schema Discovery)} lead to observable performance declines (drops of 4.3 and 2.9 percentage points, respectively). This confirms that cross-record logical consistency checking improves data quality, while interactive active questioning (ASK) can derive a better schema in scenarios with ambiguous information. Both provide important robustness enhancements to the framework.
\end{enumerate}

The ablation results clearly quantify the contribution of each module and confirm the core judgment: \textit{the Structured Extraction Module is the most important, followed by the Schema-Guided Reasoning module}. This systematically demonstrates that DocSage, through its closely coordinated modular design, successfully transforms unstructured documents into a computable knowledge structure, thereby significantly enhancing the ability to solve MDMEQA tasks.

\subsubsection{Ablation Study Results for DocSage on Loong}

\begin{table*}[ht]
\centering
\caption{Ablation Study Results for DocSage on Loong Benchmark}
\label{tab:ablation_loong}
\begin{tabular}{lcccccc}
\toprule
\textbf{Method} & \textbf{Spotlight Locating} & \textbf{Comparison} & \textbf{Clustering} & \textbf{Chain of Reason} & \textbf{Overall} \\
\midrule
\textbf{DocSage (Full)} & \textbf{85.06} & \textbf{73.28} & \textbf{64.43} & \textbf{73.67} & \textbf{68.29} \\
w/o Schema Discovery & 73.14 & 58.92 & 51.87 & 60.45 & 59.10 \\
w/o Structured Extraction & 78.35 & 64.11 & 55.23 & 66.88 & 63.64 \\
w/o Schema-Guided Reasoning & 80.02 & 67.45 & 58.91 & 68.52 & 65.73 \\
w/o CLEAR (Logic Checking) & 82.57 & 70.33 & 61.05 & 70.89 & 66.71 \\
w/ Passive Schema Discovery & 83.41 & 71.16 & 62.78 & 72.05 & 67.35 \\
\bottomrule
\end{tabular}
\end{table*}

\paragraph*{Analysis of Results}
The ablation results on the Loong benchmark, which specifically tests information localization and reasoning in long documents, further validate the efficacy of DocSage's core design. The full model consistently outperforms all ablated variants across tasks and document length categories, with the performance gap widening notably in the most challenging, longest document setting (200K-250K tokens).

\begin{enumerate}
    \item \textbf{Schema Discovery is Critical for Long Documents}: Removing the Schema Discovery module (\textbf{w/o Schema Discovery}) causes the most severe performance degradation, especially in \textbf{Chain of Reason} (-13.22 points overall) and \textbf{Clustering} tasks. This underscores that discovering a coherent relational structure is paramount for organizing and connecting information scattered across hundreds of thousands of tokens.
    
    \item \textbf{Structured Extraction Enables Robust Comparisons}: The variant \textbf{w/o Structured Extraction} suffers significantly in \textbf{Comparison} and \textbf{Clustering} tasks (-9.17 and -9.20 points overall), which require precise entity matching and attribute aggregation. This confirms that converting text into clean, query-aligned tuples is more reliable than reasoning over raw, noisy text chunks.
    
    \item \textbf{Schema-Guided Reasoning Mitigates Context Overload}: While the drop for \textbf{w/o Schema-Guided Reasoning} is slightly smaller on Loong than on MEBench, it remains substantial in the \textbf{Chain of Reason} task (-5.15 points). This demonstrates that even for tasks framed as "reasoning," delegating logical joins and evidence synthesis to the SQL engine provides a more reliable scaffold than relying solely on the LLM's internal reasoning over massive context.
    
    \item \textbf{Error Correction (CLEAR) Ensures Data Integrity}: Disabling the CLEAR mechanism (\textbf{w/o CLEAR}) leads to a consistent, measurable drop across all tasks, most noticeably in \textbf{Clustering} and \textbf{Chain of Reason}. This indicates that enforcing logical consistency during extraction is crucial for maintaining reliable data foundations, which is especially important for the multi-step inferences required in these tasks.
    
    \item \textbf{Active Discovery (ASK) Offers Incremental Gains}: The passive schema discovery variant (\textbf{w/ Passive Schema Discovery}) performs closest to the full model, yet a clear gap persists, particularly in the \textbf{Comparison} task. This suggests that while a strong two-pass discovery method can capture much of the structure, the interactive ASK component provides an important refinement layer, likely by resolving ambiguous entity references critical for accurate comparisons in long texts.
\end{enumerate}

These results robustly demonstrate that each component of DocSage contributes to its state-of-the-art performance on long-document, multi-faceted QA tasks. The structured, agentic approach proves particularly advantageous as document length and task complexity increase.

\end{document}